\documentclass[10pt,twocolumn,letterpaper]{article}

\usepackage[pagenumbers]{wacv} 

\usepackage{graphicx}
\usepackage{amsmath}
\usepackage{amssymb}
\usepackage{booktabs}
\usepackage[accsupp]{axessibility}  

\usepackage{pifont}
\usepackage{multirow}
\newcommand{\cmark}{\ding{51}}%
\newcommand{\xmark}{\ding{55}}%
\DeclareMathOperator*{\argmin}{arg\,min}

\usepackage[pagebackref,breaklinks,colorlinks]{hyperref}

\usepackage[capitalize]{cleveref}
\crefname{section}{Sec.}{Secs.}
\Crefname{section}{Section}{Sections}
\Crefname{table}{Table}{Tables}
\crefname{table}{Tab.}{Tabs.}

\def\wacvPaperID{} 
\def\confName{WACV}
\def\confYear{2024}

\begin{document}

\title{A Generative Multi-Resolution Pyramid and Normal-Conditioning 3D Cloth Draping}

\author{Hunor Laczkó$^{\ast,\dag}$\hspace{0.7cm} Meysam Madadi$^{\dag,\ddag}$\hspace{0.7cm} Sergio Escalera$^{\dag,\ddag}$\hspace{0.7cm} Jordi Gonzalez$^{\ast,\dag}$\\
{\tt\small hunor.laczko@uab.cat, mmadadi@cvc.uab.es, sescalera@ub.edu, jordi.gonzalez@uab.cat}\\
$^{\ast}$Universitat Autònoma de Barcelona, $^{\dag}$Computer Vision Center,
$^{\ddag}$ Universitat de Barcelona\\
Barcelona, Spain\\
}
\maketitle

\begin{abstract}
RGB cloth generation has been deeply studied in the related literature, however, 3D garment generation remains an open problem. In this paper, we build a conditional variational autoencoder for 3D garment generation and draping. We propose a pyramid network to add garment details progressively in a canonical space, i.e. unposing and unshaping the garments w.r.t. the body. We study conditioning the network on surface normal UV maps, as an intermediate representation, which is an easier problem to optimize than 3D coordinates. Our results on two public datasets, CLOTH3D and CAPE, show that our model is robust, controllable in terms of detail generation by the use of multi-resolution pyramids, and achieves state-of-the-art results that can highly generalize to unseen garments, poses, and shapes even when training with small amounts of data. The code can be found at: \url{https://github.com/HunorLaczko/pyramid-drape}
\end{abstract}

\section{Introduction}
\label{sec:intro}

Cloth generation and reconstruction is an active topic in computer vision with applications in virtual try-on, fashion, gaming, and movie editing, just to name a few \cite{cheng2021fashion}. 

Unfortunately, garment generation is still an open problem due to challenges such as variable topology, shape, size, style, fabric, and texture, as well as highly dynamic behavior. The problem has been tackled in the literature from different points of view: 2D image generation \cite{cui2021dressing, grigorev2021stylepeople, feng2022weakly, yang2022bodygan, jiang2022clothformer}, image-based 3D reconstruction \cite{zhao2021learning, xiu2022icon, alldieck2022photorealistic, zhu2020deep, he2021arch++, hong2021stereopifu, alldieck2019learning, zhu2022registering} and 3D garment draping and manipulation \cite{ma2021power, ma2021scale, tiwari2021neural, patel2020tailornet, bertiche2020cloth3d, bertiche2020pbns, bertiche2021deepsd,santesteban2021self,zakharkin2021point,grigorev2022hood}. This paper is focused on the latter category. 

\begin{table}[!t]
  \centering
  \small
  \begin{tabular}{@{}l@{}c@{ }c@{ }c@{ }c}
    \toprule
     & Gen. & GarmUns. & Interp. & Repr. \\
    \midrule
    SCALE \cite{ma2021scale} & \xmark & \xmark & \xmark & Patch point cloud \\
    Neural-GIF \cite{tiwari2021neural} & \xmark & \xmark & \xmark & Implicit function \\
    TailorNet \cite{patel2020tailornet} & \xmark & \xmark & \cmark & Mesh \\
    POP \cite{ma2021power} & \xmark & \xmark & \xmark & Point cloud \\
    PBNS \cite{bertiche2021pbns} & \xmark & \xmark & \cmark & Mesh \\
    Santesteban \etal \cite{santesteban2021self} & \xmark & \xmark & \xmark & Mesh \\
    DeePSD \cite{bertiche2021deepsd} & \xmark & \cmark & \xmark & Mesh \\
    Zakharkin \etal \cite{zakharkin2021point} & \xmark & \cmark & \xmark & Point cloud \\
    HOOD \cite{grigorev2022hood} & \xmark & \cmark & \cmark & Mesh \\
    CAPE \cite{ma2020learning} & \cmark & \cmark & \xmark & Mesh \\
    CLOTH3D \cite{bertiche2020cloth3d} & \cmark & \cmark & \xmark & Mesh \\
    Ours & \cmark & \cmark & \cmark & UV map \\
    \bottomrule
  \end{tabular}
  \caption{We compare our approach to state-of-the-art 3D garment generation and draping w.r.t. to their solution's properties: 1) is it Generative ({\em Gen.})?, 2) is it Garment Unspecific ({\em GarmUns.}) i.e. does not require training on many different cloth categories?, 3) does it produce Interpretable ({\em Interp.}) results?, and 4) what is the Representation ({\em Repr.})? Most of the available approaches train their specific models per each particular cloth type. Also, there are very few generative models that can output different cloth dynamics for the same input pose. Lastly, some approaches provide explainable predictions either w.r.t. local dynamics (TailorNet and ours) or physically based losses (PBNS and HOOD). Our approach covers all these points while outperforming state-of-the-art and generalizing well to unseen poses. 
}
  \label{tab:motivation}\vspace{-0.5cm}
\end{table}

\begin{figure*}[!ht]
  \centering
   \includegraphics[width=\linewidth]{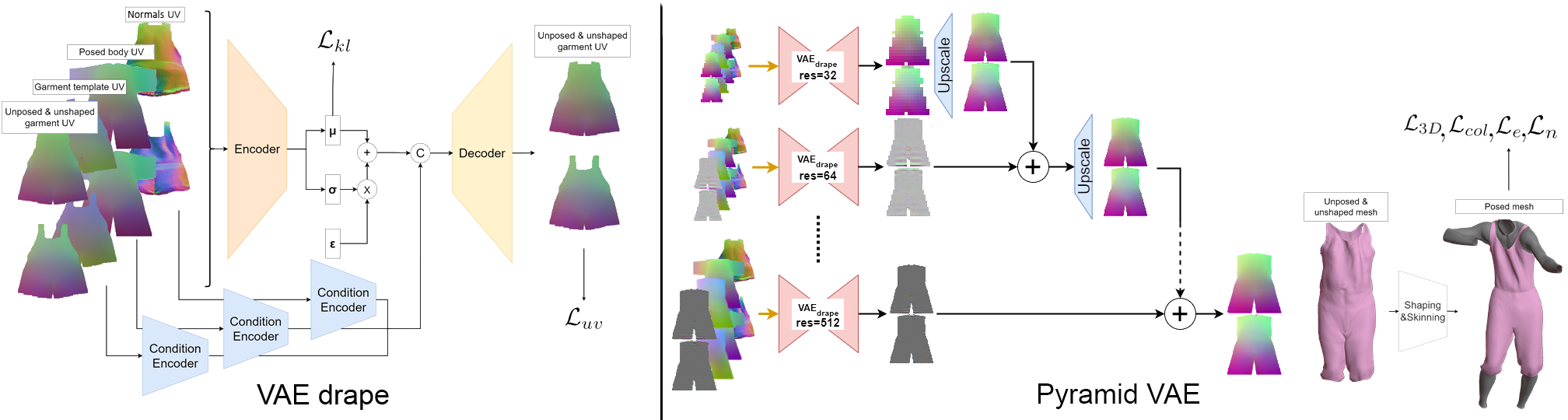}
   \caption{The proposed pyramid pipeline (right) contains basic VAE modules for each draping level (\textit{VAE}$_{drape}$, left). \textit{VAE}$_{drape}$ receives conditioning inputs and garment offsets and reconstructs the unposed and unshaped garment offsets as UV image. In the case of the first level instead of offsets, absolute coordinates are used (as shown on the left) as this will serve as a base for subsequent levels. The conditioning variables (normals, posed body, and garment template UV images) are given into three pre-trained and frozen encoders to fuse with the \textit{VAE}$_{drape}$ latent code. These conditioning encoders are trained separately in an autoencoder manner (note that normals are trained through \textit{VAE}$_{norm}$). Finally, the reconstructed UV image is converted to a mesh and passed to the skinning module after reshaping. Then, in the pyramid module, the lowest resolution level predicts low-frequency garments while the other levels are learned as offsets over their previous level. Each level output is upscaled with the proposed upscaling network and summed to the next level. At inference time, we sample from \textit{VAE}$_{norm}$ and \textit{VAE}$_{drape}$ and pass the template garment and posed body UV images.}
   \label{fig:pipeline}
\end{figure*}

Proposed solutions mainly depend on the garment representation. Meshes have been widely used to represent 3D surfaces \cite{bhatnagar2019multi, bertiche2020cloth3d,bertiche2020pbns,bertiche2021deepsd,patel2020tailornet,santesteban2021self,ma2020learning}. However, designing architectures to effectively deal with variable mesh topologies is not a trivial task. Recently, point cloud representation has received special attention thanks to the use of implicit functions \cite{gundogdu2019garnet, tiwari2021neural, hong2021stereopifu, he2021arch++, xiu2022icon} or point representations \cite{ma2021scale,ma2021power,zakharkin2021point}. Although these functions generate high-fidelity details (after converting point clouds to mesh surfaces) with an arbitrary number of output points, their high computational cost and limited quality of the generated surface when applied to unseen cases are major concerns. Lastly, UV maps representations, from texture to 3D body and cloth, have shown successful results in different application scenarios \cite{lahner2018deepwrinkles, rial2021uv, chen2022auv, su2022deepcloth, zhang2021deep, chaudhuri2021semi, xie2022temporaluv, majithia2022robust}, since they can be wrapped to a mesh at no cost while handling arbitrary topologies. We apply UV map representations in our work.

Regardless of the chosen representation, available techniques are commonly applied to (i) generate 3D garment templates in a canonical pose or (ii) drape/animate a given 3D template w.r.t. the body pose. Most state-of-the-art approaches, like ours, are based on the latter case \cite{santesteban2022snug, ma2021power, ma2021scale, tiwari2021neural, patel2020tailornet, bertiche2020cloth3d, bertiche2020pbns, bertiche2021deepsd,santesteban2021self,zakharkin2021point,bertiche2022neural}. However, there are some common limitations in most of these techniques. Firstly, cloth draping is a one-to-many problem, i.e. for a given pose and template garment there are many valid animated garments, and each garment type has different dynamical behavior. This is while these techniques overfit to the training set to predict just one single solution. Although this may be valid for some applications, it does not provide any control over the prediction. Secondly, either every single garment has its own trained model (can be categorized as supervised or unsupervised neural simulator) \cite{bertiche2021pbns,santesteban2022snug, ma2021power, ma2021scale,santesteban2021self} or one model is trained per each garment category \cite{patel2020tailornet, bhatnagar2019multi}, which limit applicability for real-world scenarios.

In order to cope with the aforementioned limitations, one of the main aims of this work is to train all sorts of garments in a \textbf{minimal} number of generative models. Generative cloth architectures have been studied before in \cite{ma2020learning,bertiche2020cloth3d} using convolutional graph networks. In this paper, we develop a variational autoencoder (VAE) on CNN architecture backbones, thanks to the use of UV maps, which have shown great results in generating plausible RGB images. Additionally, we condition the VAE on normal maps, as intermediate representations, which increases stochasticity and has been shown to be effective in gaining high-fidelity results \cite{xiu2022icon}.

Finally, there have been substantial efforts in convolutional data processing to learn features specific to different frequency domains \cite{dorta2017laplacian, saharia2022image, denton2015deep,hongPCL21}. Following this trend, we build a pyramid conditional VAE, with a series of VAEs each trained on a different image resolution. Thus, the lowest resolution VAE is responsible for learning low-frequency cloth dynamics while the other VAEs learn high-frequency details as offsets, progressively added to the previous level.

Our contributions are:
\begin{itemize}
    \item We develop a pyramid VAE for 3D cloth draping using a ConvNeXt backbone architecture \cite{liu2022convnet} and 3D surface coordinates represented as UV images.
    \item We condition the network on the surface normals unwrapped in UV maps. 
    \item We show state-of-the-art (SOTA) 3D generative cloth draping results by progressively improving the error and surface quality on two public datasets, CAPE \cite{ma2020learning} and CLOTH3D \cite{bertiche2020cloth3d}.
\end{itemize}

\begin{figure*}[!t]
  \centering
  \begin{subfigure}{0.48\linewidth}
  \centering
    \includegraphics[height=1.0in]{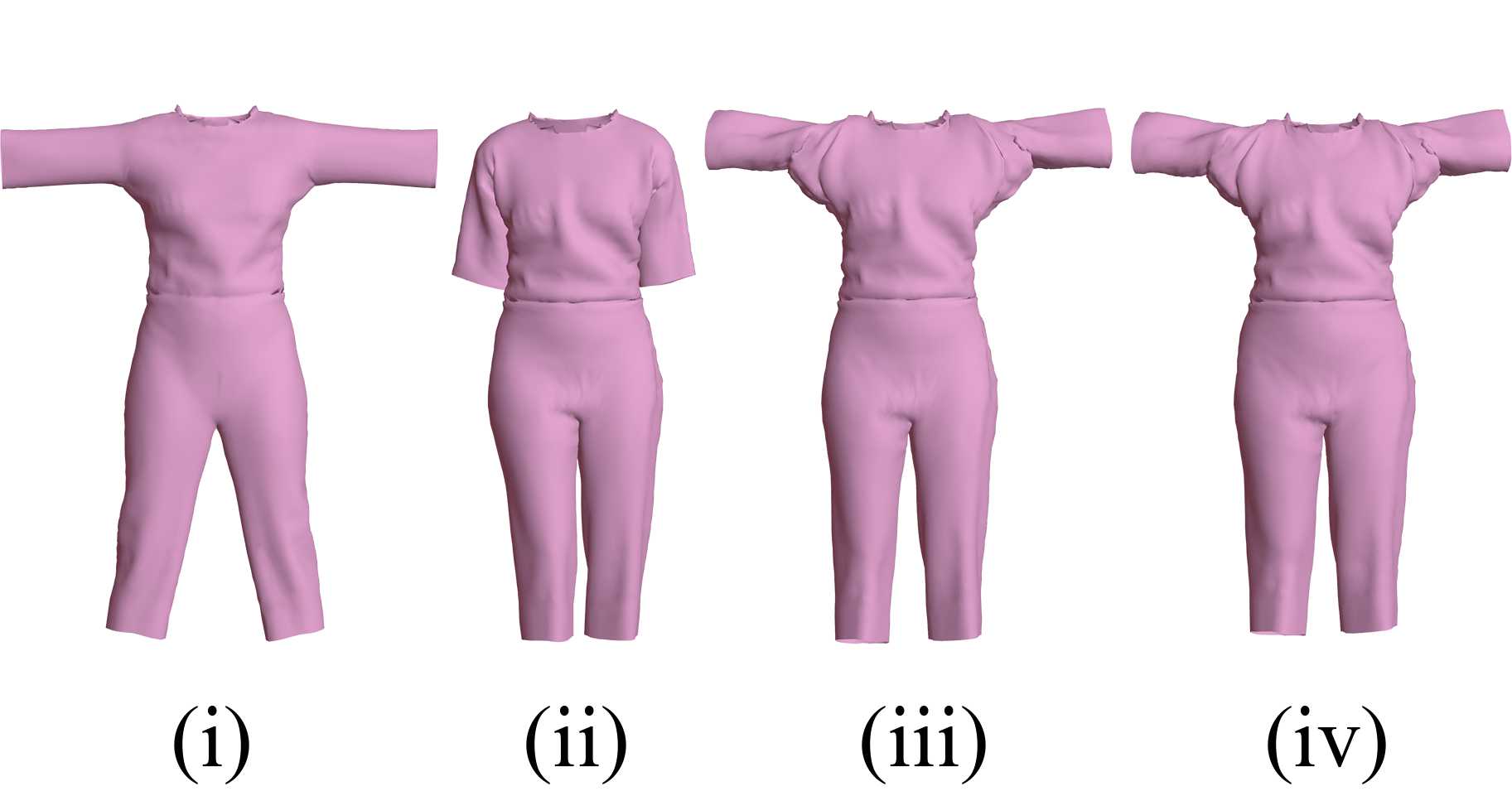}
    \label{fig:preproc-a}
    \caption{The template garment in T-pose (i) is draped on a given posed body (ii) which is unposed (iii) through an optimization process. Later it is unshaped (iv) by reversing the SMPL shaping process. Notice how the unshaped garment is wider, as the underlying body was thinner than the SMPL template.}
  \end{subfigure}
  \hfill
  \begin{subfigure}{0.48\linewidth}
  \centering
    \includegraphics[height=1.0in]{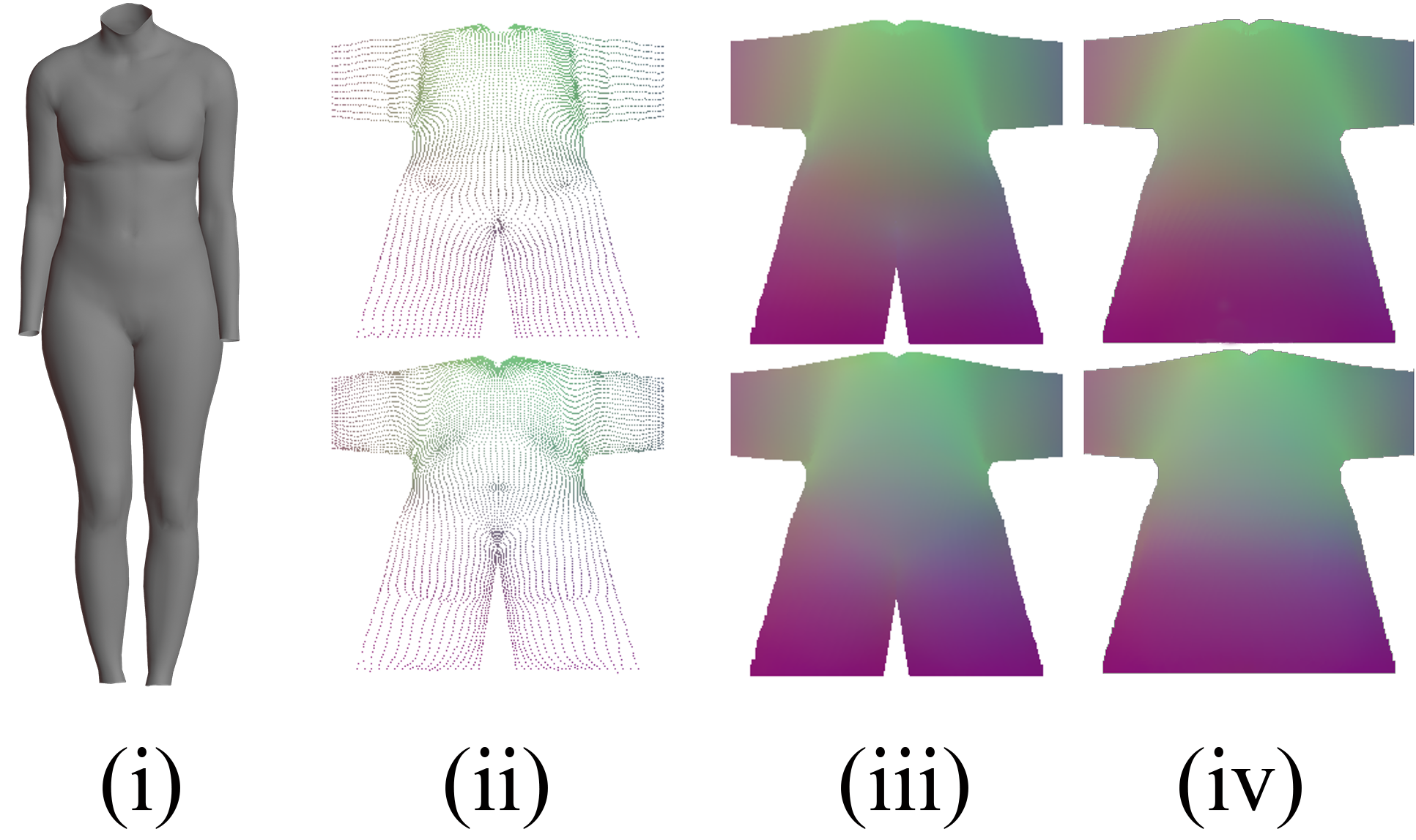}
    \label{fig:preproc-b}
    \caption{Through UV unwrapping a mapping is defined between the 3D mesh vertices (i) and the UV image pixels. After projecting the 3D mesh to the UV space the result is sparse (ii), so we inpaint the missing values by interpolation for non-skirts (iii) and skirts (iv) in two different UV templates.}
  \end{subfigure}
  \hfill
  \begin{subfigure}{0.48\linewidth}
  \centering
    \includegraphics[height=1.0in]{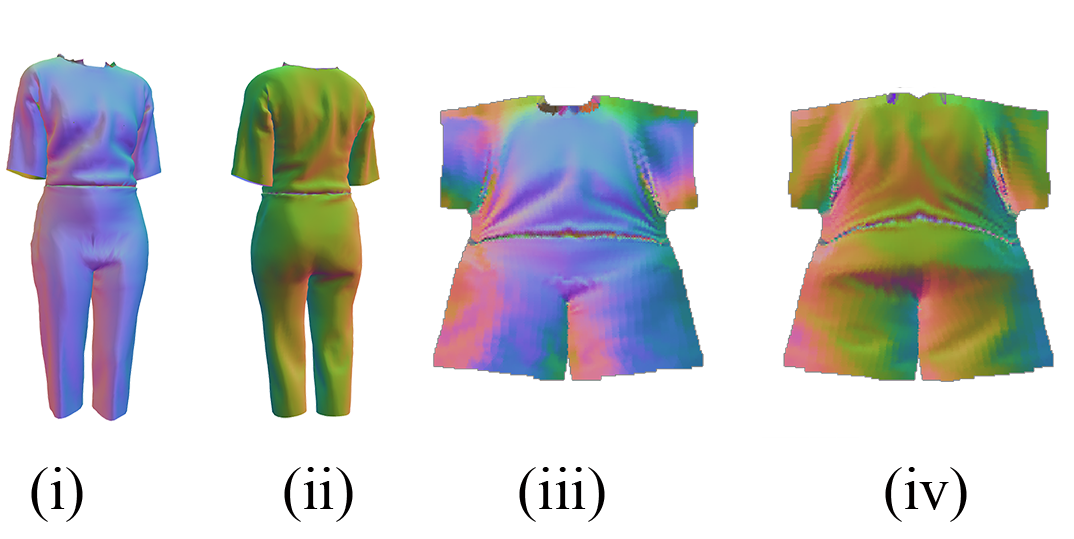}
    \label{fig:preproc-c}
    \caption{We calculate normal vectors for each vertex, visible on the front (i) and back (ii) of the mesh which are converted to front (iii) and back (iv) UV maps.}
  \end{subfigure}
  \hfill
  \begin{subfigure}{0.48\linewidth}
  \centering
    \includegraphics[height=1.0in]{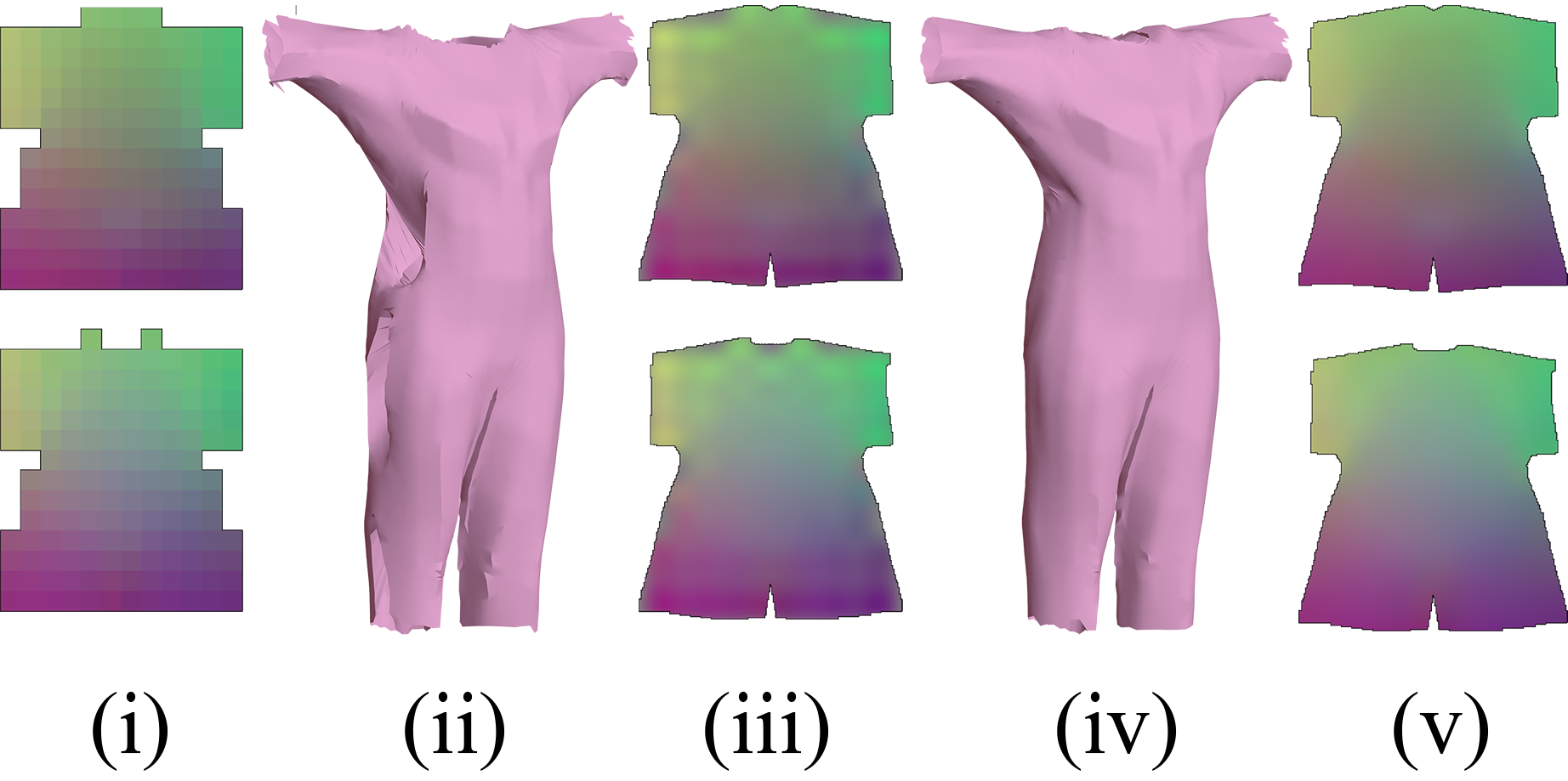}
    \label{fig:preproc-d}
    \caption{Upscaling from a low-resolution UV image (i) results in artifacts around the boundaries (ii-iii) using regular upscaling. By the proposed upscaling (iv-v) this can be avoided. }
  \end{subfigure}
  \caption{Examples of preprocessed data. a) Cloth unposing and unshaping. b) 3D mesh to UV map. c) Surface normals calculation. d) UV image down/upscaling.}
  \label{fig:preproc}\vspace{-0.3cm}
\end{figure*}

\section{Related works}
We compare our proposed architecture to the state-of-the-art in \cref{tab:motivation}. Most of the available works are deterministic approaches \cite{ma2021scale,tiwari2021neural,patel2020tailornet,ma2021power,bertiche2021pbns,santesteban2021self,bertiche2021deepsd,zakharkin2021point,grigorev2022hood} in which a single solution is predicted for the given inputs. There has been a set of approaches \cite{ma2021scale,tiwari2021neural,ma2021power,bertiche2021pbns,santesteban2021self}, among deterministic works, in which a single garment instance is trained over many given body poses, called neural simulation, depicting cloth simulation in graphic engines. The works of \cite{ma2021scale,ma2021power,sanchez2021physxnet} convert the given body surface into a UV map and learn 2D body pose features through CNN architectures, similar to us. In \cite{ma2021power} an additional garment geometry feature tensor is learned through auto-decoding. Later 2D sampled features along with positional encodings are given to a shared MLP to predict an offset vector added to the sampled body point. Authors in \cite{tiwari2021neural} use a combination of several MLPs as an implicit function to first unpose sampled points and add offset vectors to them. The updated points along with pose encodings are given to a signed distance function to predict the point occupancy. Learning features in a canonical pose has been a common technique in state-of-the-art. Similar to us, \cite{santesteban2021self} preprocess the garments to form a collision-free unposed and unshaped garment. Later a VAE is trained to learn garment-specific deformations. Although they use a VAE for training, their approach is not designed to be generative at inference time. Rather, they predict the encoded features from the body shape parameters and motion information. In a different solution, TailorNet \cite{patel2020tailornet} decomposes the garment into high and low-frequency surfaces in a canonical pose. Then the wrinkles are formed by a mixture of high frequencies. All the aforementioned approaches were trained supervisedly. PBNS \cite{bertiche2020pbns} is the first unsupervised approach that uses physically based losses to train a lightweight neural network. They apply an MLP to extract features from the pose which is multiplied to an optimizable garment geometry matrix. The outputs are offsets added to the template garment in a canonical pose. Later, this type of solution has been extended to include temporal information \cite{santesteban2022snug,bertiche2022neural}.

Among deterministic approaches, several works develop garment unspecific models in which a single model is trained to handle all sorts of garments. DeePSD \cite{bertiche2021deepsd} is inspired by the physically-based simulation formulation and can handle variable topologies during training without registration. They train the network by a combination of supervised and unsupervised losses. Zakharkin \etal \cite{zakharkin2021point} transforms the posed body point cloud into the final garment by conditioning on the garment latent code. HOOD \cite{grigorev2022hood} develops a new message-passing mechanism in graph neural networks and trains the network with self-supervised physically based losses. 

To the best of our knowledge, there are a few generative methodologies in 3D garment draping \cite{ma2020learning,bertiche2020cloth3d}. These works apply VAE graph neural networks on registered meshes in a canonical pose. CLOTH3D \cite{bertiche2020cloth3d} proposes a new topology for the body to model skirts-like clothes. CAPE \cite{ma2020learning} trains the network by an additional adversarial loss. Our approach decomposes the surface into different frequencies and learns the details progressively showing accurate reconstruction capabilities.

\section{Generative 3D cloth draping}
\label{sec:method}

The proposed 3D cloth draping pipeline is: 1) generative and stochastic, 2) efficient, 3) general, as it does not require separate models for each garment type, and 4) able to handle both high and low-frequency details. We assume a) SMPL body pose and shape parameters \cite{loper2015smpl} plus a garment template in canonical pose (T-pose for example) are available for the draping, b) the garment size is matched with the body shape\footnote{Although garment resizing is possible as a preprocessing to match the body shape, we consider it as out-of-the-scope of this paper.}, c) the garment has a low elasticity, thus the mesh edge lengths should not change drastically after the draping, and d) the garment has a single surface layer (\eg pockets are not considered). Also, we apply our experiments to daily life clothes. Next, we explain the details and architectural choices to fulfill the aforementioned goals. The architecture pipeline is shown in \cref{fig:pipeline}.

\subsection{Preprocessing}
\label{sec:preproc}

\subsubsection{Cloth unposing and unshaping}
We follow the trends in the literature \cite{ma2020learning,ma2021power,ma2021scale,patel2020tailornet,bertiche2020cloth3d} that force the network to make predictions in a canonical pose space to improve the results. We explore this idea in several aspects: 1) eliminating the global rotation, 2) unposing, and 3) unshaping. Let $\theta\in\mathbb{R}^{24\times3\times3}$ and $\beta\in\mathbb{R}^{1\times10}$ be the SMPL pose and shape parameters, $B\in\mathbb{R}^{10\times(m\times3)}$ and $W\in\mathbb{R}^{m\times24}$ the SMPL blend shape and blend weights, $D\in\mathbb{R}^{m\times3}$ the shaped and posed SMPL body vertices, $G_t\in\mathbb{R}^{n\times3}$ a template garment with $n$ vertices registered on SMPL mesh (see next section for details) and $M\in\{0,1\}^m$ its registration mask (note $n=\sum M$), and $G\in\mathbb{R}^{n\times3}$ the $G_t$ garment draped on $D$. To eliminate the global orientation, we first update the garment and body vertices by $G\cdot\theta^{0}$ and $D\cdot\theta^{0}$, and the $\theta$ root joint rotation ($\theta^0$) by an identity matrix. Then, we transform $G$ to a canonical pose space by:
\begin{equation}
    G_c = \argmin_{G_c} \| \mathcal{S}(G_c+[\beta \cdot B]^M,\theta; W) - G \|,
    \label{eq:unposing}
\end{equation}
where $\mathcal{S}$ is the skinning function and $[.]^M$ is the indexing based on mask $M$. An example can be seen in Fig. \ref{fig:preproc}(a).

\subsubsection{Cloth registration}
We follow CLOTH3D \cite{bertiche2020cloth3d} protocol for registration, that is 1) increasing SMPL resolution and removing head, hands, and feet vertices ($m=14475$ after this update), and 2) creating an extra body mesh for skirt-like garments. Additionally, we apply laplacian smoothing to the body and template garment and use a non-rigid ICP to warp the template garment to the body surface this way achieving a less noisy registration. The mask $M$ is extracted by taking the nearest body vertices. Finally, we compute the registration as follows.

The simplest registration solution is by taking body vertices and finding the nearest vertex on the warped garment. However, it is possible that a garment vertex is assigned to two or more body vertices. This causes null faces on the garment which degrades the garment resolution and leads to noisy loss functions. Instead of the nearest mesh vertex, we find the nearest point on the garment surface through a fast numerical approximation. To do so, we generate a random vector $\omega^{k\times3}$ for each garment face in which $1=\sum_{j=1}^3 \omega_{ij}$ where $i\in\{1..k\}$. We use $\omega$ as a weighting over the face vertices coordinates to sample points on each garment face. Finally, we find the nearest points from this randomly sampled set and save the $\omega$ values for the nearest point and its corresponding face index. This is used to reconstruct the garment in any frame represented by SMPL topology.

\subsubsection{3D mesh to UV map}
UV map representation has been used by several approaches \cite{lahner2018deepwrinkles, rial2021uv, chen2022auv, su2022deepcloth, zhang2021deep, chaudhuri2021semi, xie2022temporaluv}. Its advantage over meshes is that it is a 2D image, which makes it applicable in history-rich GPU-efficient CNN models. However, a UV map is only partially continuous, unlike a mesh. The reason is that to unwrap a 3D mesh into 2D, the surface is torn into several segments to fit in. To minimize the number of segments and distortions, we follow \cite{madadi2020learning} in which seamlines at the sides of the body are manually defined where they divide the body into two segments: front and back. Based on this idea we create UV images with a $512\times256$ resolution. In the supplementary material, we explain the details of how we create the UV pixel coordinates corresponding to 3D vertices to have a 3D reconstructed surface. An example can be seen in Fig. \ref{fig:preproc}(b). Through this operation we convert $D$, $G$, $G_t$ and $G_c$ into $D^{uv}$, $G^{uv}$, $G_t^{uv}$ and $G_c^{uv}$, respectively. Also, the mask $M$ is updated to $M^{uv}$ accordingly.

Note that although vertex indices are identical between skirt-like and trousers-like clothes after registration, they do not follow the body in the same way and also have different dynamic behaviors. Therefore, a single template for UV coordinates for both of them is not valid. For this reason, we group garments into skirts and non-skirts, create two UV templates for each, and train them independently. This allows us to cover all garment types minimally with only two models, unlike other methods that require a separate model for each type.

\subsubsection{Surface normals}
Surface frequency details can be partially reflected in the surface normals as the direction of the normals and their relative change compared to their neighborhood correlate to the details of the surface. 
An example can be seen in Fig. \ref{fig:preproc}(c). As the figure shows, small wrinkles and even large deformations are easy to identify. This helps guide the model to be able to recreate these details.

\subsubsection{UV image down/upscaling}
\label{sec:upscale}

Regular RGB images can be rescaled by simple bilinear or bicubic algorithms. Applying these techniques to UV images will result in invalid values and artifacts near the boundary areas (reflecting in the seamline between the two sides of the mesh) due to zero background pixels. Fortunately, downsampling can be done by a fast nearest neighbor interpolation to fill the background. However, it is not accurate for upscaling. To overcome this, we develop a fast and differentiable interpolation block by utilizing a CNN with five convolutional layers only to fill the background. For training, we use MS-SSIM loss on the already available data from the previously downscaled images. Later, to train the pyramid VAE, we freeze this upscaling network. We apply image scaling by a factor of 2 and have UV resolutions in the range $64\times32$ and $512\times256$. The results can be seen in Fig. \ref{fig:preproc}(d).

\subsection{Conditional pyramid VAE}
\label{sec:arch}
In the basic definition of cloth draping, a garment is draped on the body given a pose. However, this problem has unlimited valid solutions. Therefore, models with a single solution do not truly formulate the problem. Generative models are able to deal with this issue, though sometimes they suffer from not being stochastic and become one-to-one functions. Cloths are highly dynamic objects and some approaches either converge to smooth and average dynamics \cite{bertiche2020cloth3d,ma2020learning} or overfit to specific categories and poses \cite{patel2020tailornet}. To cope with these issues, we build our generative model based on a VAE (called \textit{VAE}$_{drape}$) that is conditioned on the garment template $G_t^{uv}$, posed body $D^{uv}$ and normal map $N^{uv}$, and reconstructs $\hat{G}_c^{uv}$\footnote{$\hat{.}$ means estimated by the network.}. Then, the 3D garment $\hat{G}$ can be obtained by:
\begin{equation}
    \hat{G} = \mathcal{S}([\hat{G}_c^{uv}]^{J\odot M}+[\beta \cdot B]^M,\theta, W)
\end{equation}
where $J$ contains corresponding UV indices to SMPL vertices masked by $M$ and $\odot$ is element-wise product.

Although the normal map is not available at inference time, its generation and sampling is a much easier problem than 3D cloth draping. Therefore, we build a second VAE for normals (called \textit{VAE}$_{norm}$) that is again conditioned on $G_t^{uv}$ and $D^{uv}$, and reconstructs $\hat{N}^{uv}$. Due to its simplicity, adaptation, and performance, we adapt ConvNeXt architecture \cite{liu2022convnet} as the VAEs' encoders,  thanks to the usage of UV images. Decoders are mirrors of encoders (downsampling convolutional layers are replaced with transpose convolutions). We explain the adapted architecture and training details in the supplementary material.

As a common problem for CNNs in 3D space, low-frequency dynamics are learned first and high-frequency details are ignored. To solve this issue, we propose a pyramid network that starts from a low-resolution image to learn the global shape of the cloth. Then, the surface quality is progressively improved in the next higher resolution levels as offsets to the previous level. Note that we use the idea in Sec. \ref{sec:upscale} to upscale images between consecutive levels. See \cref{fig:pipeline} for architecture details.

\subsection{Losses}
\label{sec:loss}
We use the following loss functions to train the network:
\begin{itemize}
    \vspace{-0.2cm}
    \item L2 loss between $\hat{G}_c^{uv}$ and $G_c^{uv}$ masked by $M^{uv}$ ($\mathcal{L}_{uv}$).
    \vspace{-0.6cm}
    \item KL loss ($\mathcal{L}_{kl}$).
    \vspace{-0.2cm}
    \item L2 loss between $\hat{G}$ and $G$ ($\mathcal{L}_{3D}$).
    \vspace{-0.2cm}
    \item Garment against body collision loss ($\mathcal{L}_{col}$) \cite{bertiche2021deepsd}: 
    \begin{equation}
        \mathcal{L}_{col} = \sum_{(i,j)\in \mathcal{N}} min(\mathbf{v}_{j,i}\cdot \mathbf{N}_j - \epsilon, 0)^2,
        \label{eq:collision}
    \end{equation}
    where $\mathcal{N}$ is the set of nearest neighbor correspondences between the predicted cloth and body, $\mathbf{v}_{j,i}$ is the connecting vector between the $j$-th body vertex to the $i$-th cloth vertex, $\mathbf{N}_j$ is the $j$-th body vertex normal, and $\epsilon=4$mm is a small threshold. This loss works well when the cloth is close to the body and pushes back collided vertices back to the nearest body vertex. 
    \vspace{-0.6cm}
    \item L1 loss on the mesh edges lengths between $\hat{G}$ and $G$ ($\mathcal{L}_e$).
    \vspace{-0.2cm}
    \item L1 loss on 3D surface normals between $\hat{G}$ and $G$ ($\mathcal{L}_n$).
\end{itemize}
Then, we define the regularization, reconstruction, and final losses as:
\begin{equation}
\mathcal{L}_{reg} = \mathcal{L}_{col} + \mathcal{L}_{e} + \mathcal{L}_{n} + \mathcal{L}_{3D},
\end{equation}
\begin{equation}
\mathcal{L}_{rec} = \mathcal{L}_{uv} + \lambda \cdot \mathcal{L}_{reg},
\end{equation}
\begin{equation}
\mathcal{L} = \mathcal{L}_{rec} + \delta \cdot \mathcal{L}_{kl},
\end{equation}
where $\lambda$ is a coefficient controlling the regularization losses and $\delta$ is a balancing term between the reconstruction loss and the KL divergence. Tuning $\delta$ is important to preserve the stochasticity of the VAE while still reaching a low reconstruction error. Therefore, we use $\beta$-VAE technique \cite{burgess2018understanding} to train the network. We start the training by setting $\delta<10^{-3}$ and once we observe a flat $\mathcal{L}_{rec}$ we linearly increase $\delta$. 

\section{Experiments}
\label{sec:exp}

\textbf{Datasets.} We use two common 3D human bodies and garments datasets CLOTH3D \cite{bertiche2020cloth3d} and CAPE \cite{ma2020learning} to train our model and compare against SOTA. First, we use CLOTH3D, a large-scale synthetic dataset of clothed humans. It contains more than two million frames from several thousand sequences with a large variety of garment types, topologies, shapes, and motions. We use 20 percent of the original training dataset to speed up the training and evaluate our method on the defined test set.

Second, we use CAPE, a smaller dataset created with high-resolution 3D scans of real subjects. As it is smaller, it captures a smaller variety of garment types and subjects but still covers a wide range of motions. The scans are registered on top of the SMPL body and encoded as offsets over the minimally clothed body. Unlike CLOTH3D, the bodies and garments are available as a single entity and the model needs to predict them as a single mesh. The dataset is split into two separate datasets for female and male subjects, both containing four cloth types. Each of these datasets contains around 20-30K frames for training and around 5K for testing.

\textbf{Metrics.} Since we have SMPL body-garment registration available for both datasets, we are able to calculate a simple point-to-point Euclidean distance averaged across the garment. Therefore, we follow SOTA protocol to report the \textit{reconstruction error}\footnote{Note that to do so, we use the full VAE model including encoders and decoders.} with two common metrics: the average per vertex Euclidean error (V2V) and the Chamfer distance (CD) in mm. Following SOTA, the reported errors are the averages over the average per garment type. We also calculate additional qualitative metrics: 1) the percentage of collided vertices and 2) the absolute difference of the surface area between the draped and template garments in $m^2$. In the second metric, we analyze the amount of stretching or compression imposed by the model on the input template garment. While small differences are expected, large changes indicate that the prediction does not conform to the input template.

\subsection{Ablation results}
\label{sec:ablation}

We perform an ablation of our approach on the CLOTH3D dataset.

\begin{table}[]\centering\footnotesize
\begin{tabular}{{@{}llccccc@{}}}
\toprule
Data & Experiment & Pyr. 1 & Pyr. 2 & Pyr. 3 & Pyr. 4 \\
\midrule
\multicolumn{6}{l}{\textbf{Non-skirt} (baseline: 7.11)} \\
\midrule
 & Incremental training& 11.97 & 7.62 & 7.20 & 7.14 \\
 & End-to-end w/o reg. & 10.07 & 5.41 & 4.33 & 3.79 \\
 & End-to-end w/ reg.  & \textbf{9.68} & \textbf{5.30} & \textbf{4.10} & \textbf{3.78} \\
 \midrule
\multicolumn{6}{l}{\textbf{Skirt} (baseline: 9.79)} \\
\midrule
 & Incremental training& 15.78 & 9.49 & 9.13 & 8.90 \\
 & End-to-end w/o reg. & 12.00 & 6.74 & 5.38 & 4.60 \\
 & End-to-end w/ reg.  & \textbf{11.94} & \textbf{6.60} & \textbf{5.00} & \textbf{4.28} \\
 \midrule
\multicolumn{6}{l}{\textbf{Combined} (baseline: 8.01)} \\
\midrule
 & Incremental training& 13.24 & 8.24 & 7.84 & 7.72 \\
 & End-to-end w/o reg. & 10.72 & 5.85 & 4.68 & 4.05 \\
 & End-to-end w/ reg.  & \textbf{10.43} & \textbf{5.73} & \textbf{4.40} & \textbf{3.95} \\
\bottomrule
\end{tabular}
\caption{Detailed V2V results of the ablation study on CLOTH3D dataset. We can see the effect of each contribution across each level of the pyramid. It is also compared to the baseline, non-pyramid model. Note that the combined error is not the average of the non-skirt and skirt numbers, but the average of the six garment categories combined.}
\label{tab:ablation}
\end{table}

\begin{table}[]\centering\footnotesize
\begin{tabular}{{@{}lcccccc@{}}}
\toprule
 & \multicolumn{2}{c}{$\mathcal{L}_{e}$} & \multicolumn{2}{c}{$\mathcal{L}_{n}$} & \multicolumn{2}{c}{Collision (\%)} \\
 \midrule
 & Non-sk. & Skirt & Non-sk. & Skirt & Non-sk. & Skirt \\
 \midrule
Baseline      & 0.580 & 1.326 & 0.150 & 0.161 & 5.49 & 2.64 \\
Increm.       & 0.578 & 1.581 & 0.159 & 0.184 & 5.45 & 3.10 \\
E2e w/o reg.  & 0.706 & 1.058 & 0.126 & 0.133 & 2.00 & 0.63 \\
E2e w/ reg.   & \textbf{0.447} & \textbf{0.906} & \textbf{0.124} & \textbf{0.132} & \textbf{1.98} & \textbf{0.61} \\
\bottomrule
\end{tabular}
\caption{Regularization loss values for the ablation experiments}
\label{tab:regularization_losses}\vspace{-0.3cm}
\end{table}

\subsubsection{Contributions}
We specifically create and train four different models to validate our contributions:
\begin{itemize}
    \item \textbf{Baseline}: A single level \textit{VAE}$_{drape}$ at full resolution ($512\times256$).
    \vspace{-0.2cm}
    \item \textbf{Incremental}: Pyramid architecture built from several \textit{VAE}$_{drape}$ submodels at different resolutions. Trained incrementally, one level at a time. In this approach, we start by training the lowest resolution level, then freeze it, add the next level, and train again. We repeat until we train all levels.
    \vspace{-0.2cm}
    \item \textbf{End-to-end}: Final pyramid model where all levels were trained at the same time. The condition encoders are still frozen. See details in the supplementary material.
    \vspace{-0.2cm}
    \item \textbf{End-to-end with regularization}: Final pyramid model where all levels were trained at the same time with the regularization losses ($\mathcal{L}_{reg}$ in \cref{sec:loss}) applied during training. A $\lambda=0.1$ was used.
\end{itemize}

\begin{figure}[!ht]
  \centering
   \includegraphics[width=1\linewidth]{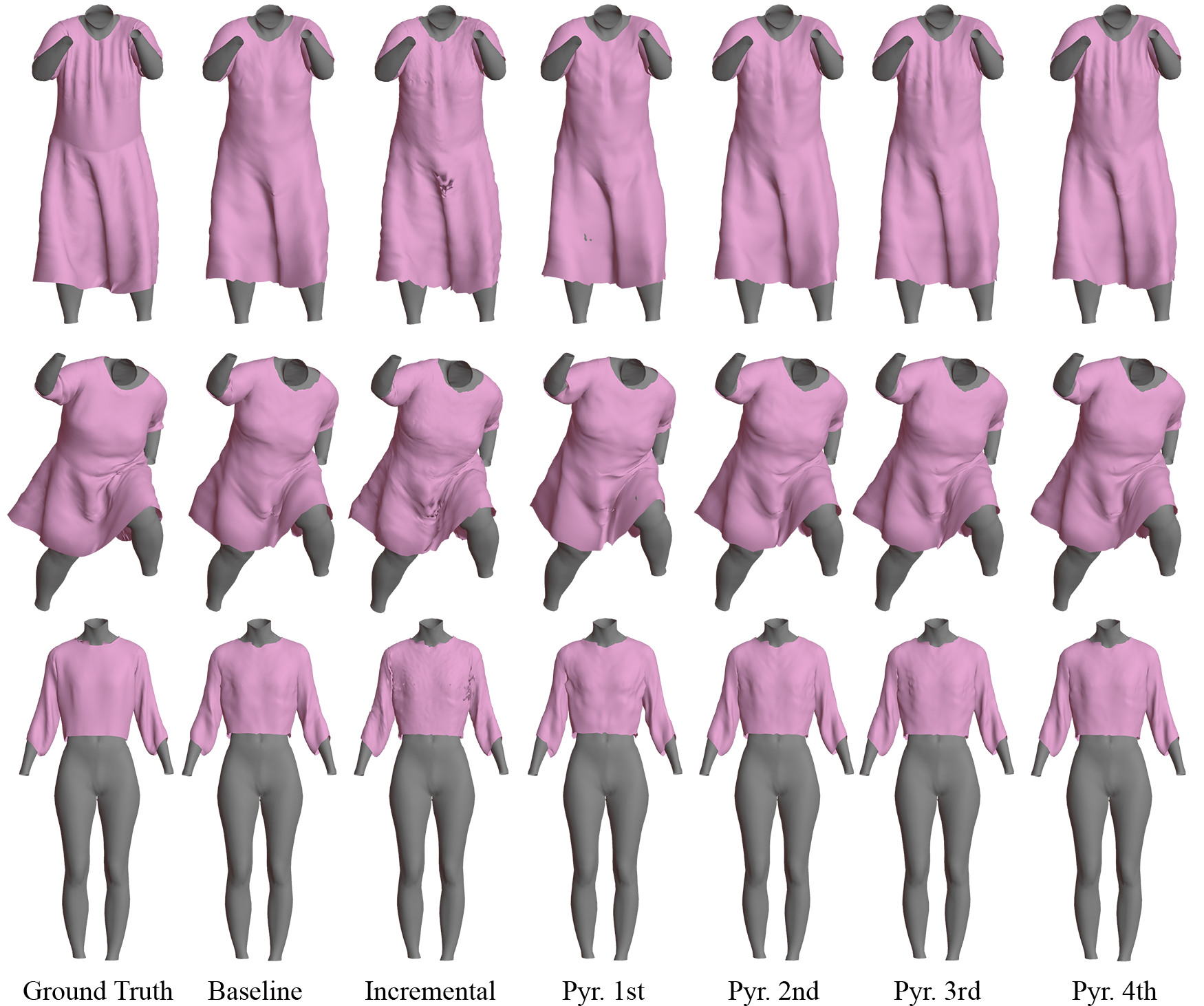}
   \caption{Effectiveness of the pyramid architecture on three examples. Ground truth, with baseline and last output of the incremental method, followed by the output of the final pyramid after each level after skinning and shaping. Notice that each subsequent layer adds details to the previous output.}
   \label{fig:ablation_pyramid}\vspace{-0.3cm}
\end{figure}

We compare the contributions quantitatively and qualitatively in \cref{tab:ablation} and \cref{fig:ablation_pyramid}, respectively. The baseline results are comparable with the pyramid network when trained incrementally. In \cref{fig:ablation_pyramid} it can be seen that the baseline produces good results, but lacks finer details. 

Next, we see the output of the incremental method. The aim of this approach is to provide a stable training process and more control over the individual levels. In this case, higher levels do not affect the prediction of lower levels. From \cref{fig:ablation_pyramid}, we can see that while this approach achieves a low error, the outputs present some noise along with some artifacts. Since there is no feedback from higher levels if there is an error in one of the lower levels, this needs to be corrected later, which is not always possible. 

Furthermore, we present our end-to-end trained model, where each pyramid level was trained simultaneously. By adding supervision at the output of each level, we can still have stable training, where each level outputs a valid prediction. In this approach, there is information flowing both ways between levels. As such, the higher levels can help the lower levels produce more accurate details which in turn lowers the amount of corrections needed to be learned in higher levels. Overall, end-to-end training helps to improve the results by a large margin over the incremental training. \cref{fig:ablation_pyramid} shows that each new level does not only add more details to the garment, but it can also fix small errors or artifacts present in lower levels.

We also explore the use of different regularization losses during training as presented in \cref{sec:loss}. As we can see in \cref{tab:ablation}, these losses help to improve the results while they have a higher impact on skirts compared to non-skirts. We further analyze the impact of these losses in \cref{tab:regularization_losses}. It can be seen that adding the regularization decreases the edge loss and the percentage of the cloth vertices colliding with the body. A lower value on the edge loss means that the model is more robust against compression and stretching of the topology of the mesh.

\subsubsection{Sampling}

First, we sample the normal maps from \textit{VAE}$_{norm}$ as these are not available at inference time. Further details about \textit{VAE}$_{norm}$ can be found in the supplementary material. 

\begin{figure}[!ht]
  \centering
   \includegraphics[width=0.8\linewidth]{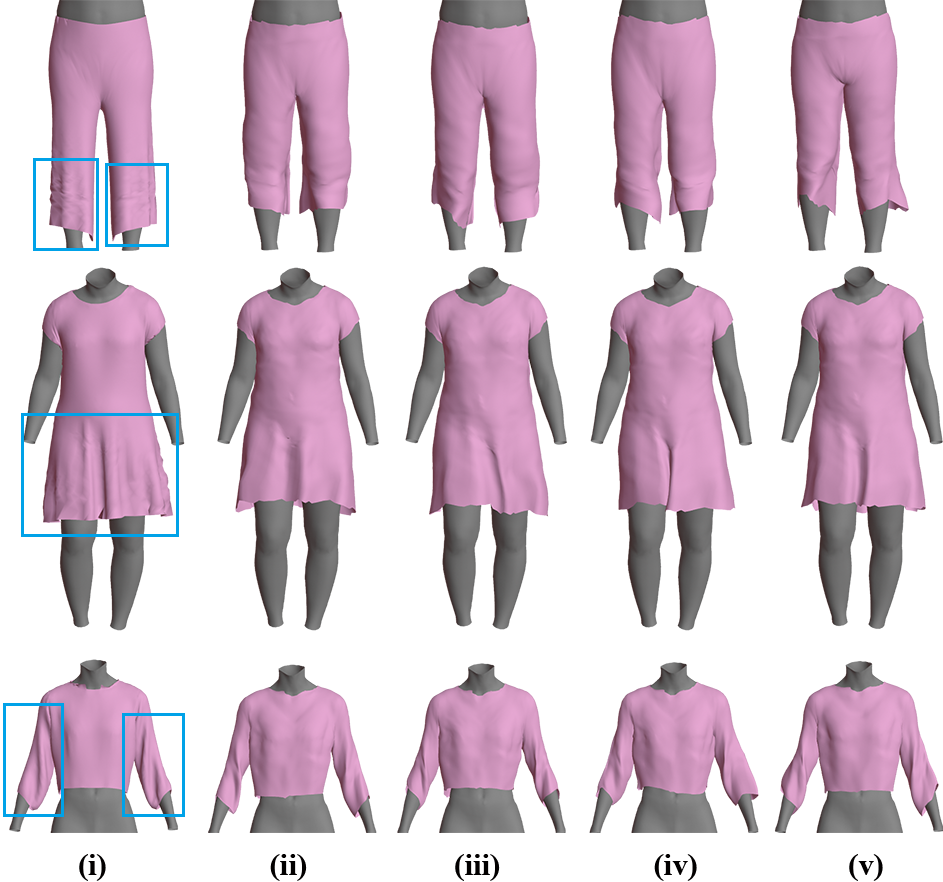}
   \caption{Sampling capabilities of the model. (i) ground truth, (ii-v) results by sampling the VAE latent spaces. Notice the produced changes in the blue rectangles. The shown examples have an average area difference of less than 8\% compared to the ground truth. }
   \label{fig:vae_sampling}\vspace{-0.3cm}
\end{figure}

After sampling normal maps, we can also study how they can help \textit{VAE}$_{drape}$ in the generation of drapings dealing with the one-to-many assignment problem. By fixing the template garment and body pose, we randomly sample from \textit{VAE}$_{norm}$ and \textit{VAE}$_{drape}$ and generate garment UV maps. The results in \cref{fig:vae_sampling} show diverse dynamics in different garment parts. We can also see that the surface areas of the samples only vary around 8\%. These minor variations can be attributed to the stretching or compression of the garment. Therefore, the resulting garments negligibly differ from their template shape.

\subsection{SOTA comparison}
\label{sec:sota}
\begin{figure}[!ht]
  \centering
   \includegraphics[width=\linewidth]{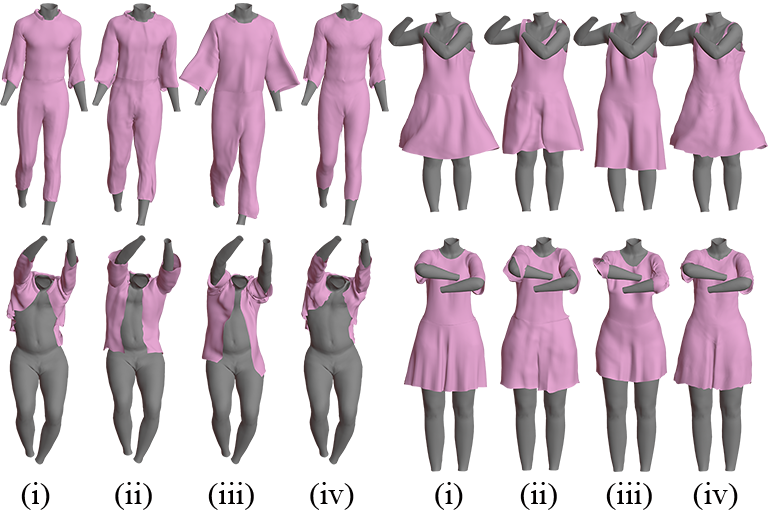}
   \caption{Qualitative comparison of ground truth (i) with predictions from DeePSD (ii), HOOD (iii) and our method (iv). HOOD shows improvement over DeePSD in the level of detail. However, it over-stretches the garment. Our method offers a high level of detail while having the minimum stretching and compression over the template shape.}
   \label{fig:hood}\vspace{-0.3cm}
\end{figure}

We compare our final contribution to generative (CLOTH3D) and non-generative (DeePSD and DeepCloth) SOTA on the CLOTH3D dataset in \cref{tab:cloth3d_sota}. As can be seen, our methodology outperforms the SOTA by a large margin (79.46\% relative improvement). We also use the qualitative metrics introduced at the beginning of \cref{sec:exp} to compare our approach with HOOD and DeePSD. We follow the instructions in their GitHub pages and train these methods on the same CLOTH3D subset we used to train our model. As it can be seen in \cref{tab:hood}, we achieve a small area difference similar to DeePSD, while HOOD over-stretches the garment to favor the collision. DeepSD has the worst collision among others. HOOD shows the worst CD error. However, given that it is a self-supervised model, comparing it to the ground truth is not fair. We qualitatively compare with SOTA in \cref{fig:hood}. Although HOOD shows realistic draping, it drastically changes the template shape. In our experiments with HOOD, changing the elasticity hyperparameter made very small differences. DeePSD shows the least level of detail among others. Additionally, we observe an over-emphasized presence of physical forces, like gravity, in both HOOD and DeePSD. Overall, our method shows better qualitative results than SOTA.

\begin{figure}[t!]
  \centering
   \includegraphics[width=0.9\linewidth]{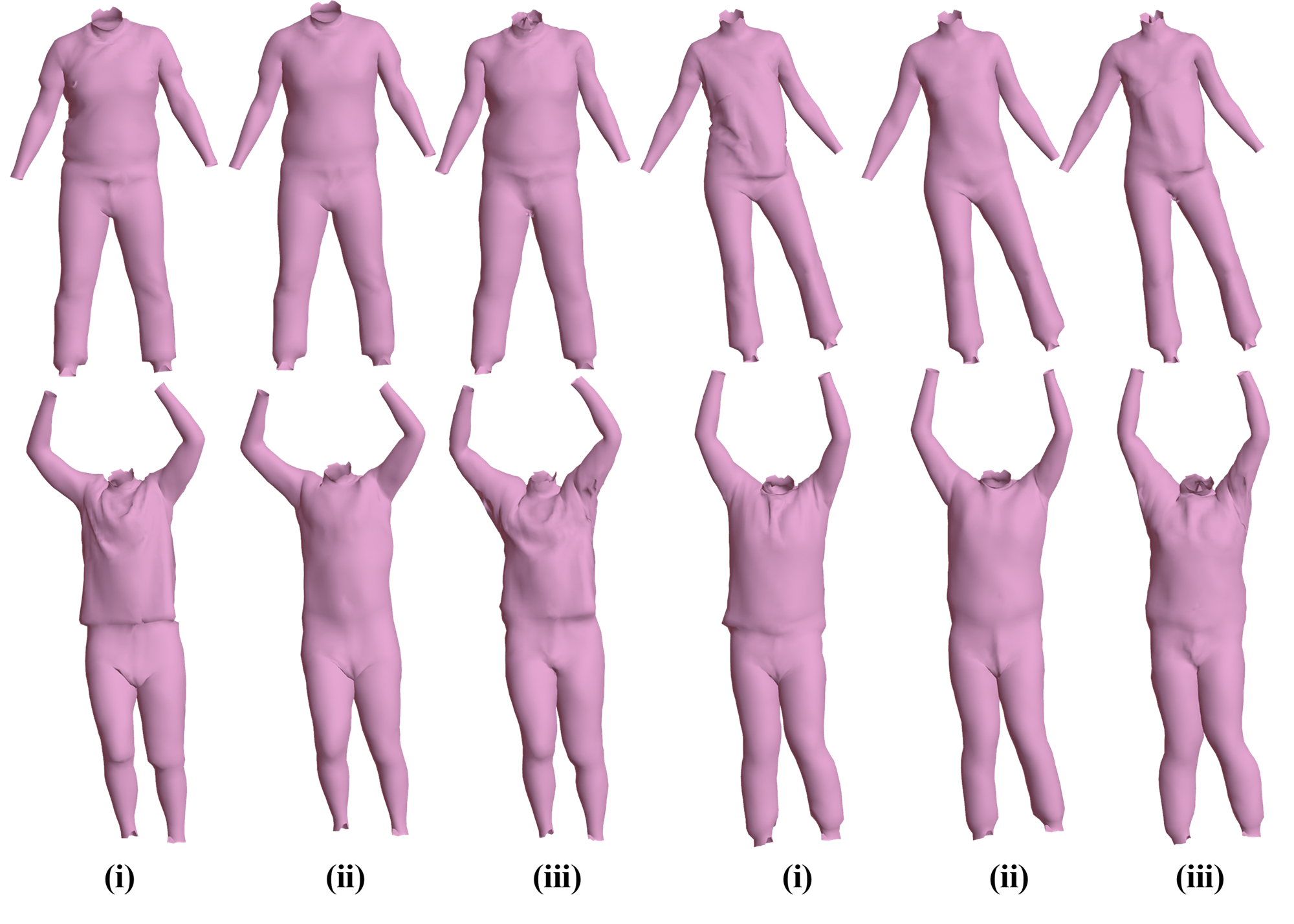}
   \caption{Qualitative examples from CAPE dataset comparing ground truth (i) to CAPE (ii) and ours (iii).}
   \label{fig:cape_qual}\vspace{-0.3cm}
\end{figure}

We also compare our method to CAPE, a generative SOTA, on their dataset. As one can see in \cref{tab:cape_sota}, our pyramid model progressively improves the results and we achieve a relative improvement of 40.5\% on the male split, and 13.9\% on the female split of the CAPE dataset at the third level of the pyramid. We also notice some overfitting towards the last level of the pyramid architecture. This can be attributed to the fact that the meshes are lower resolution compared to CLOTH3D, and they contain fewer details. Qualitative results in \cref{fig:cape_qual} show how our approach reconstructs more cloth details while CAPE resembles the body.

\begin{table}[]
\begin{minipage}[l]{0.48\linewidth}
\centering
\begin{tabular}{{@{}lc@{}}}
\toprule
Method                & V2V      \\
                      & error \\
\midrule
CLOTH3D \cite{bertiche2020cloth3d}             & 29.00     \\
DeePSD \cite{bertiche2021deepsd}               & 23.78     \\
DeepCloth \cite{su2022deepcloth}               & 19.23     \\
Ours                                           & \textbf{3.95} \\
\bottomrule
\end{tabular}
\captionof{table}{SOTA comparison on CLOTH3D dataset. We show large performance improvements over SOTA.}
\label{tab:cloth3d_sota}
\end{minipage}%
\hfill%
\begin{minipage}[c]{0.48\linewidth}
\vspace{-0.3cm}
\centering
\begin{tabular}{{@{}lcc@{}}}
\toprule
Method                & Female        & Male          \\ 
\midrule
Pyr. 1 lvl & 6.63          & 8.94          \\
Pyr. 2 lvl & 3.92          & 5.44          \\
Pyr. 3 lvl & \textbf{3.11} & \textbf{3.66}          \\
Pyr. 4 lvl & 3.36          & 4.00   \\
CAPE \cite{ma2020learning} & 3.61          & 6.15          \\
\bottomrule
\end{tabular}
\captionof{table}{SOTA comparison on CAPE dataset (V2V error). CAPE \cite{ma2020learning} numbers are taken from the authors' GitHub page.}
\label{tab:cape_sota}\vspace{-0.3cm}
\end{minipage}\vspace{-0.3cm}
\end{table}

\begin{table}[]
\centering
\begin{tabular}{{@{}lccc@{}}}
\toprule
Method                & CD error   & Collision (\%) & Area diff  \\
\midrule
DeePSD \cite{bertiche2021deepsd} & 36.85 & 3.29 & 0.05429 \\
HOOD \cite{grigorev2022hood}    & 42.99  & \textbf{0.44} & 0.13211  \\
Ours                            & \textbf{11.59} & 1.04 &  \textbf{0.05282} \\
\bottomrule
\end{tabular}
\caption{Qualitative metrics comparing with SOTA on CLOTH3D dataset.}
\label{tab:hood}\vspace{-0.3cm}
\end{table}

\subsection{Performance}
Thanks to the convolutional architecture and having relatively light models, we can achieve fast inference times. The whole pyramid architecture has 40M parameters and without VAE encoders it decreases to 26M. Using an Nvidia 3090 GPU, we can achieve an inference time (for sampling and generation without encoders) of 50ms and 60ms using a batch size of 1 and 4, respectively. This allows near real-time inference.

\section{Conclusions}
\label{sec:conclusion}

We presented an effective approach to solve the problem of 3D garment draping and generation. By projecting the 3D meshes to UV images we were able to leverage established convolutional methods to create a pyramid VAE architecture. We observed how different pyramid levels allow the model to add different levels of detail to the generated cloth meshes. We further conditioned the network on normal maps, which provide additional regularization for the network to better focus on the details. Our results show SOTA performance on two public datasets, CAPE and CLOTH3D. We further showed that we can generalize to several garment types with a minimal number of models. 

\textbf{Limitations and future work.} There are some limitations of our work that remain interesting for future research. Our current approach requires garment registration on top of the SMPL model as a preprocessing. The results also depend on the quality of the registration. Our model does not handle multi-layer garments. Finally, although we obtained SOTA results by using default SMPL blend weights, learning garment, and pose-dependent blend weights may provide extra improvements to surface quality.

\section{Acknowledgement}
This work has been partially supported by the Spanish projects PID2022-136436NB-I00, PDC2022-133305-I00, TED2021-131317B-I00, by ICREA under the ICREA Academia programme. The authors also acknowledge the support of the Spanish Ministry of Economy and Competitiveness (MINECO) and the  European Regional Development Fund (ERDF) under Project No. PID2020-120611RB-I00/AEI/10.13039/501100011033.

{\small
\bibliographystyle{ieee_fullname}
\bibliography{egbib}
}

\clearpage
\renewcommand*{\thesection}{\Alph{section}}
\renewcommand*{\thefigure}{\Alph{figure}}
\renewcommand*{\thetable}{\Alph{table}}
\setcounter{section}{0}
\setcounter{table}{0}
\setcounter{figure}{0}

\twocolumn[{%
\renewcommand\twocolumn[1][]{#1}%
\begin{center}
      {
      \Large \bf A Generative Multi-Resolution Pyramid and Normal-Conditioning 3D Cloth Draping \\ Supplementary material      \par
      \vspace*{24pt}
      \lineskip .5em
      \par
      }
      \vspace*{12pt}
   \end{center}
}]

In this supplementary material, we provide further details on the architecture (\cref{sec:architecture}), training (\cref{sec:training}), building template UV coordinates (\cref{sec:uv}), and additional qualitative results (\cref{sec:results}).

\section{Architecture\protect\footnote{We commit to releasing the code upon the paper's acceptance.}}
\label{sec:architecture}

\subsection{Baseline}
As we showed in Fig. 1 in the main paper, \textit{VAE}$_{drape}$ is a conditional VAE based on ConvNext \cite{liu2022convnet}. ConvNext is made up of stages, where after each stage the image or its representation is downscaled. Each stage can have a different depth, meaning the number of blocks it contains. We empirically update the number of blocks and features at each level. In this case, the depths are 3, 3, 3, 9, and 3 with 32, 64, 128, 256, and 128 $(=F)$ features, respectively. The baseline receives an input of resolution $512\times256$ and the encoder ConvNext has five stages, this way the output features will have a size $8\times4\times F$. The condition encoders work similarly. The dimensionality $F$ is set to 128, 16, 32, and 80 for the \textit{VAE}$_{drape}$ latent vector, template garment, body, and normals condition encoders, respectively. The combined dimensionality of condition encoders is 128 to balance between conditioning features and \textit{VAE}$_{drape}$ latent vector. We also use a tanh activation on the conditioning latent codes to ensure we have the same scale in the feature space. Finally, all the latent codes are concatenated before feeding the decoder. For the decoder, we mirror the encoder architecture and replace the downsampling step with upsampling using transposed convolutions. The output size is the original $512\times256$ resolution. We use tanh activation on the last decoder layer.

\subsection{Conditioning autoencoders}
\label{sec:cond}
We explained conditioning encoders in the baseline architecture. However, these encoders are part of their own autoencoders, pretrained and frozen. That means one autoencoder is used to train each of the three conditioning inputs: template garment, posed body, and normal map. The encoder parts of these networks will act as the condition encoders. When incorporating them in the main pipeline, their weights are frozen to ensure that the quality of the features remains intact. We use tanh activation on the last decoder layer.

The conditioning network for the normal map is the \textit{VAE}$_{norm}$. The reason for this design is that normals are not available at inference time. Therefore, we create a generative pipeline for them such that we can sample from it at inference time. \textit{VAE}$_{norm}$ itself is conditioned on the template garment and posed body UV images. This is because we want the normal features to be disentangled from the other conditioning variables.

\begin{figure}[!t]
   \centering
   \includegraphics[width=\linewidth]{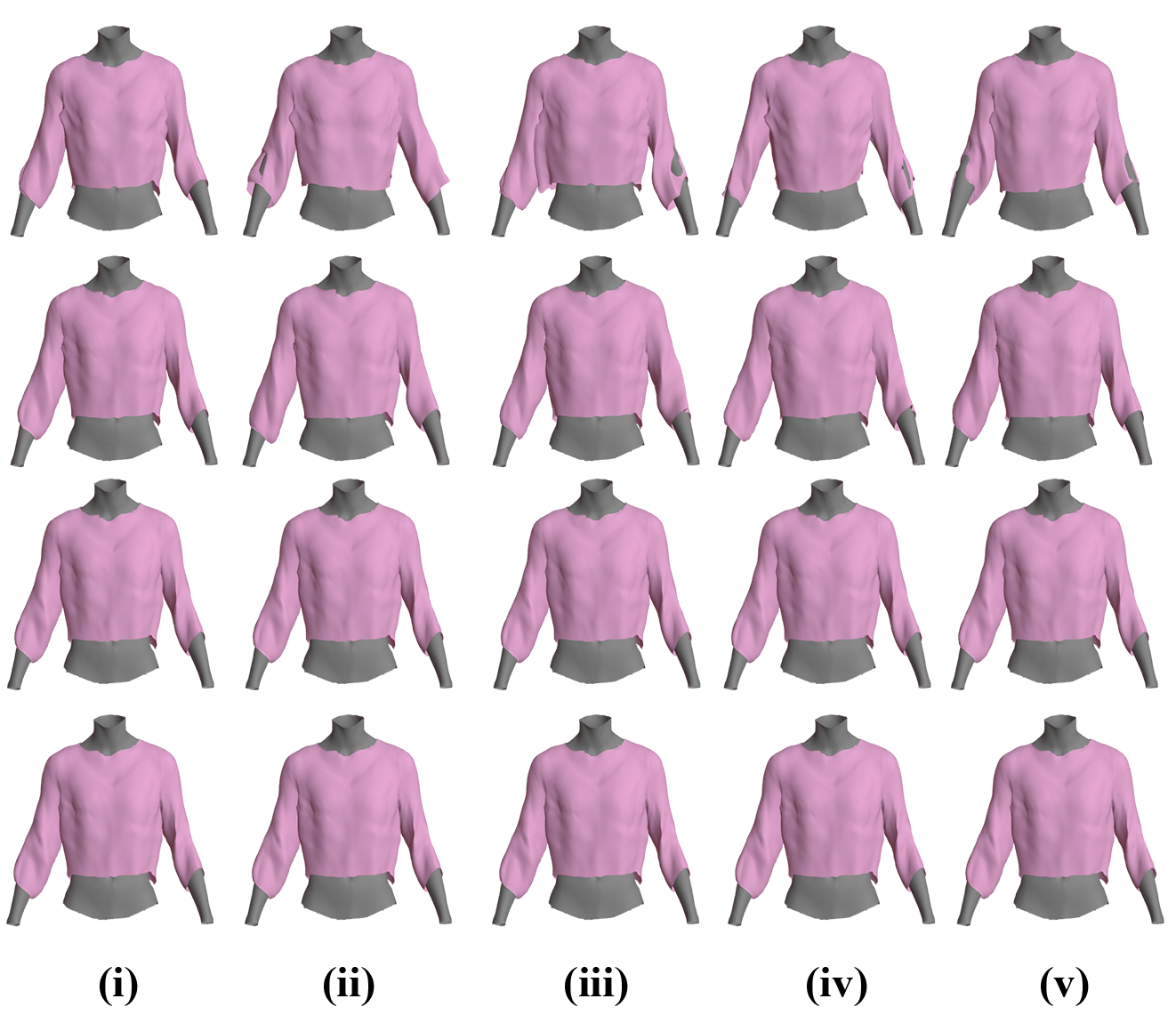}
   \caption{We show the pyramid sampling at each level, one level per row. The first element of each row is the base for the next row.}
   \label{fig:sampling_steps}
\end{figure}

\begin{figure}[!t]
   \centering
   \includegraphics[height=19cm]{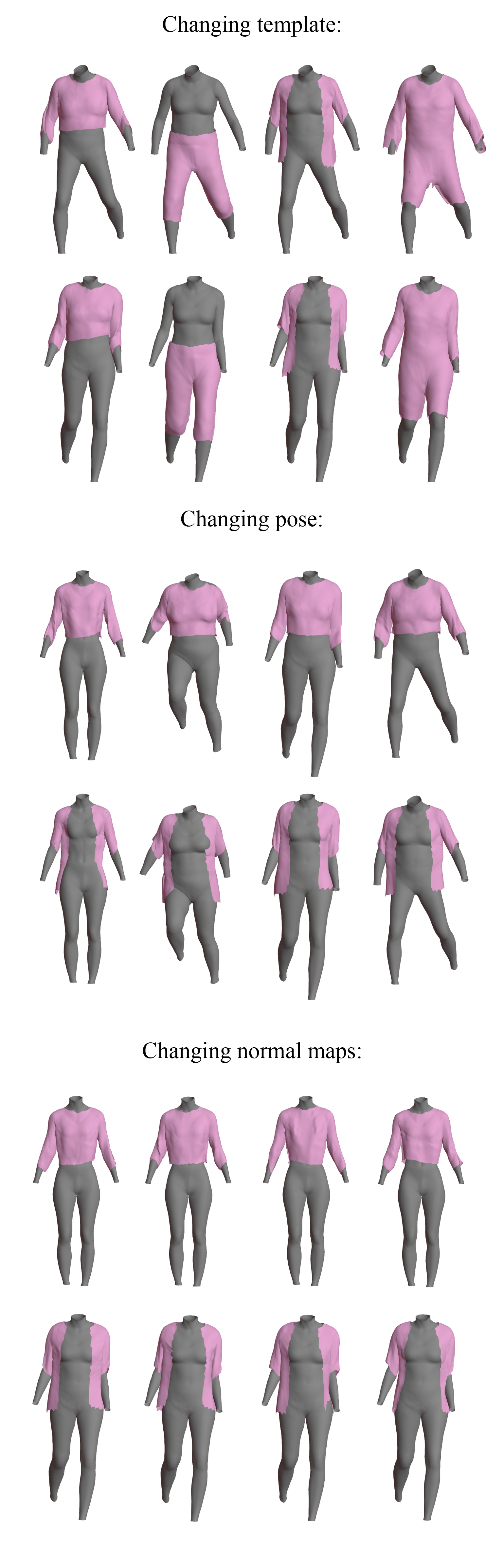}
   \caption{We show the sampling process when changing only one condition. In each case, the stated condition is changing, and the other two and the latent code are fixed. For each case we have two examples as the two rows.}
   \label{fig:mix_match}
\end{figure}

\subsection{Pyramid}
The pyramid model is built using several \textit{VAE}$_{drape}$ models of varying image resolutions. The main difference between them is the number of stages, which have to be decreased for lower levels of the pyramid. As such we use 5, 4, 3, and 2 stages for the resolutions $512\times$256, $256\times128$, $128\times64$, and $64\times32$, respectively. Similarly, we decrease the kernel sizes of the convolutions and use kernel sizes of 7, 5, 5, and 3 for the mentioned resolutions. Note that we keep the same feature dimension of $8\times4\times128$ for all the levels' latent codes.

The first level provides the base for the prediction which is a low-resolution UV image and all subsequent layers are added on top of this as offsets. To achieve this, the first level receives the low-resolution ground truth as input, while later layers receive the offset between the ground truth and the previous level's output along the conditioning inputs. The output's resolution matches that of the input. In order to be able to add the offsets from the next level, this output needs to be upscaled. For this, we use a custom model to upscale the UV images (see sec. \ref{sec:upscale_suppl}). Repeating this at each level, we double the resolution after each level until we reach the final $512\times256$ resolution.

\subsection{Upscaling model}
\label{sec:upscale_suppl}
As we mentioned in the main paper we design an upscaling model to deal with the values on the garment mask boundaries within the pyramid network. This model contains a network of 5 convolutional layers with relu activation, batch normalization, and 256, 128, 64, 32, and 3 feature sizes, respectively, and the kernel size follows the corresponding pyramid-level kernel size. The last convolution has a tanh activation instead of relu to predict the 3D coordinates. The upscaling procedure is as follows. First, the image is passed to the above network. This outputs an image that is multiplied with the background mask (inverse of the garment mask) and added to the input image. Finally, we resize this combined image using bilinear interpolation.

\section{Training}
\label{sec:training}

Due to the number of modules to be trained, we first apply incremental training in which each sub-module is trained independently. Later we will try end-to-end training of the levels (conditioning encoders are still frozen). Next, we will present these sub-modules individually.

\subsection{Conditiong autoencoders}
The initial set of modules encodes the conditional inputs of the pipeline (as explained in sec. \ref{sec:cond}). We train these networks once on the highest resolution. We use Multiscale-SSIM \cite{zwang03} loss for the autoencoders, and add an extra KL loss (using a $\beta$-VAE strategy) when training \textit{VAE}$_{norm}$. Finally, we optimize the networks with a batch size of 4 and Adam optimizer with a learning rate $10^{-4}$.

After the training, we are able to achieve a reconstruction error of 1.1mm with the posed body and 9.3mm with the template garment autoencoders. Note that the available data for the template garments is only 1.6 thousand outfits which are further split into training and validation data. As for the \textit{VAE}$_{norm}$, we achieve an MS-SSIM value of 0.993 with an L1 error of 0.04.

\begin{figure}[t!]
   \centering
   \includegraphics[width=\linewidth]{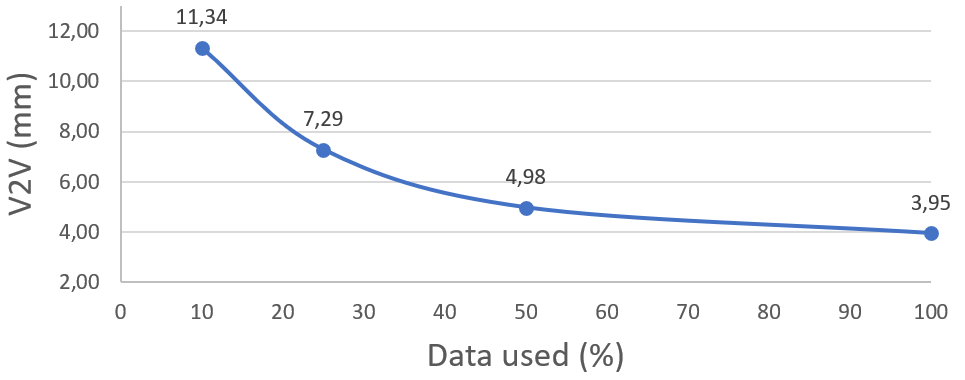}
   \caption{Generalization capability of our pyramid network on CLOTH3D dataset as a function of the amount of training data used.}
   \label{fig:generalization_trend}
\end{figure}

\subsection{Pyramid}
In the pyramid pipeline, each level is trained individually in the incremental approach and together in the end-to-end approach. The advantages of both cases are detailed in the main paper Sec. 3.1.1. During training, each level receives the same conditional features generated from the high-resolution conditional input, thereby avoiding the need to train three encoders per level and generate separate conditions for each level. This has significant speed improvements. For the variational encoders, each condition and the unposed garment are downscaled to the given resolution. The decoder outputs an image at the same resolution which is upscaled to be added to the next level's output.

The losses used are described in detail in the main paper. Note that we only start using the normal loss after 2000 iterations, since otherwise, it makes the initial steps of the training highly unstable.

We optimize the network with a batch size of 4 and Adam optimizer with a learning rate $10^{-4}$.

\section{Template UV coordinates}
\label{sec:uv}

A template UV map is obtained by projecting a three-dimensional body mesh to a fixed two-dimensional shape. To construct this mapping, we start by annotating the body mesh vertices according to the front and back of the body. This is followed by defining some keypoints: 1) around the ends of legs and arms, 2) around the neck, and 3) bilaterally for knees, hips, and underarms. We also create a low-resolution 2D mesh as a template for the UV map. We define pairs of previously selected key points and corresponding points on the UV map template. Using these control points we apply thin plate spline interpolation (TPS) to morph the 3D vertices to the 2D template (a 3D flat surface) once for the front and then for the back vertices. The vertices along the border between the front and back belong to both parts, so they have to be duplicated. During the 3D reconstruction, these duplicated pixels will be averaged to obtain the 3D coordinate.

To further improve the construction of UV maps, we define more keypoints automatically. We take the vertices of the previously mentioned border and find the closest vertex on the 2D template. This vertex is obtained by calculating distances between the points and also from their normal vectors, the latter ensuring that the mesh is stretched outwards from the center to the template. With these new keypoint pairs, we apply TPS again. Finally, we obtain UV mapping that maximizes the data representation by covering a large portion of the given rectangular grid (as shown in Fig. 2(b) in the main paper).

\section{Results}
\label{sec:results}

\subsection{Pyramid sampling}

We demonstrate how the sampling process works in the pyramid architecture. Given a set of conditioning variables, which are the template garment, posed body, and normal maps, we sample the VAE latent space, concatenate the encoded conditions, and pass it to the decoder. When we do this at the first pyramid level, we get some potential base garments as seen in \cref{fig:sampling_steps}. After this, we sample from the second level and generate the offsets over the first level, which adds some smaller details. The contribution of each level can be seen in the figure. In the case shown, after the second level, the changes are minimal.
To further demonstrate the sampling possibilities, we show examples of only one of the conditions changing to better demonstrate how it works. This can be seen in \cref{fig:mix_match}.

\subsection{Normal map sampling}

\begin{figure}[ht!]
   \centering
   \includegraphics[width=\linewidth]{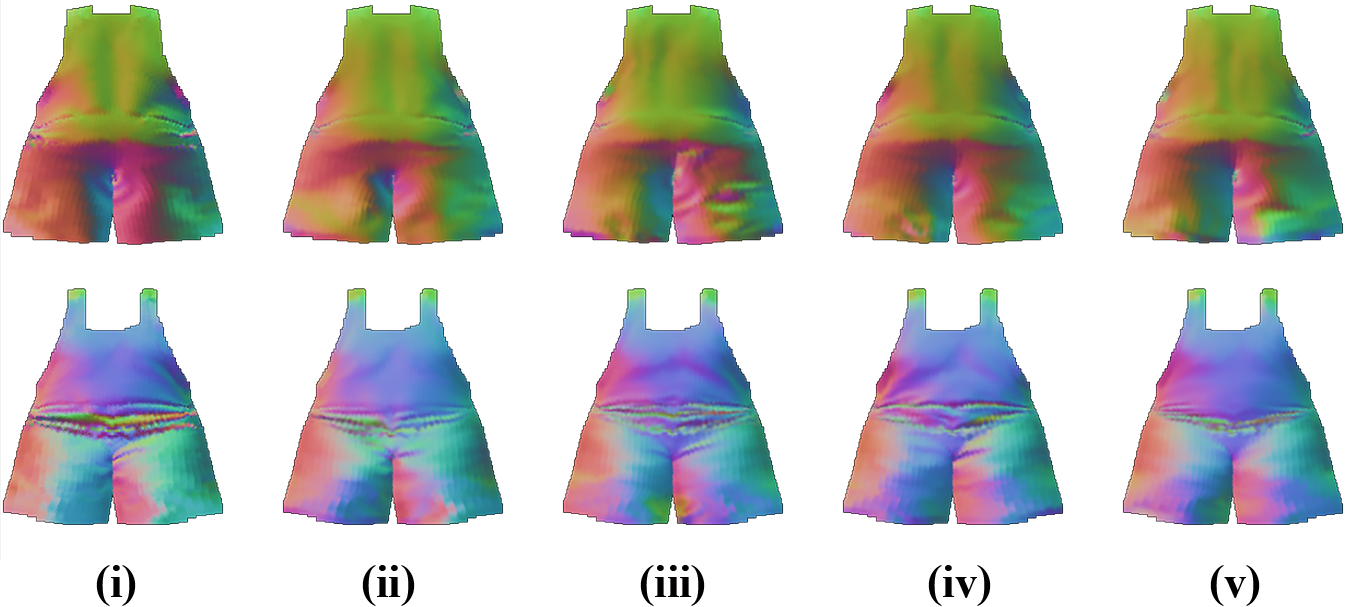}
   \caption{Examples of sampling normal maps. The left (i) is the original, and the four on the right (ii-v) are randomly sampled from the \textit{VAE}$_{norm}$ model. We can see high variance while they still represent realistic deformations.}
   \label{fig:normal_sampling}
\end{figure}

The aim of the normals is to provide guidance and variability for the generation of samples. We study the \textit{VAE}$_{norm}$ model's ability to generate normal maps. We can see some examples in \cref{fig:normal_sampling} which show how we can generate multiple plausible normal maps for a given input.

\subsection{Generalization}
To show the generalization and representation capacity of our model we evaluate how the error scales when using a smaller portion of the training data. We run the training with the following portions of the data: 0.5, 0.25, and 0.1\footnote{Note these portions are not applied on the whole CLOTH3D dataset but the 20\% we initially selected in this paper.}. As seen in the error trend in \cref{fig:generalization_trend}, we can achieve state-of-the-art with only 10 percent of the data. It also shows we can scale the method by increasing the amount of data, but the improvements will diminish soon. 

\subsection{Additional example results}
\begin{figure*}[]
  \centering
   \includegraphics[width=0.95\linewidth]{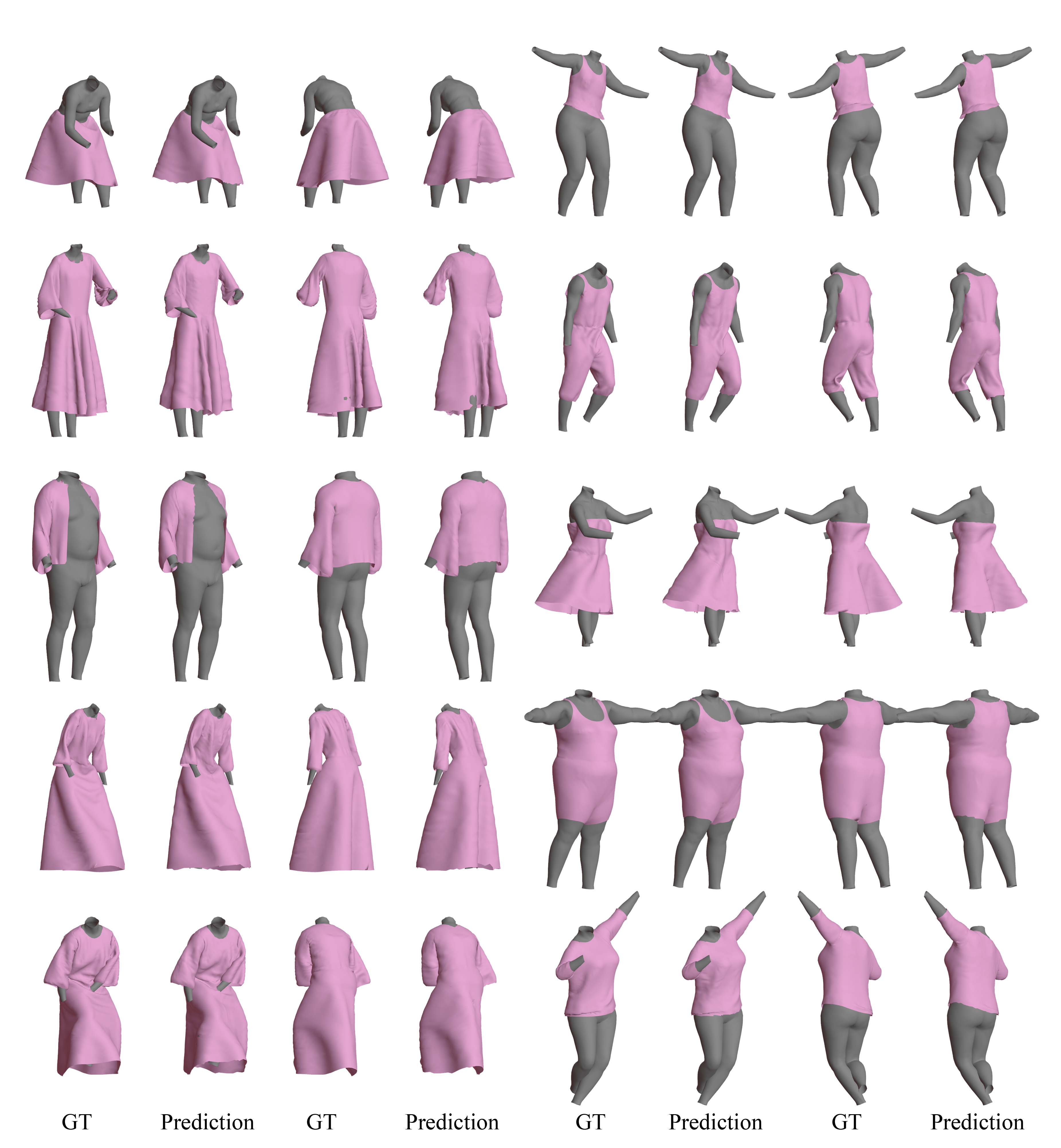}
   \caption{Additional qualitative examples from the CLOTH3D dataset: ground truth and proposed model's reconstruction.}
   \label{fig:examples_1}
\end{figure*}

\begin{figure*}[]
  \centering
   \includegraphics[width=0.95\linewidth]{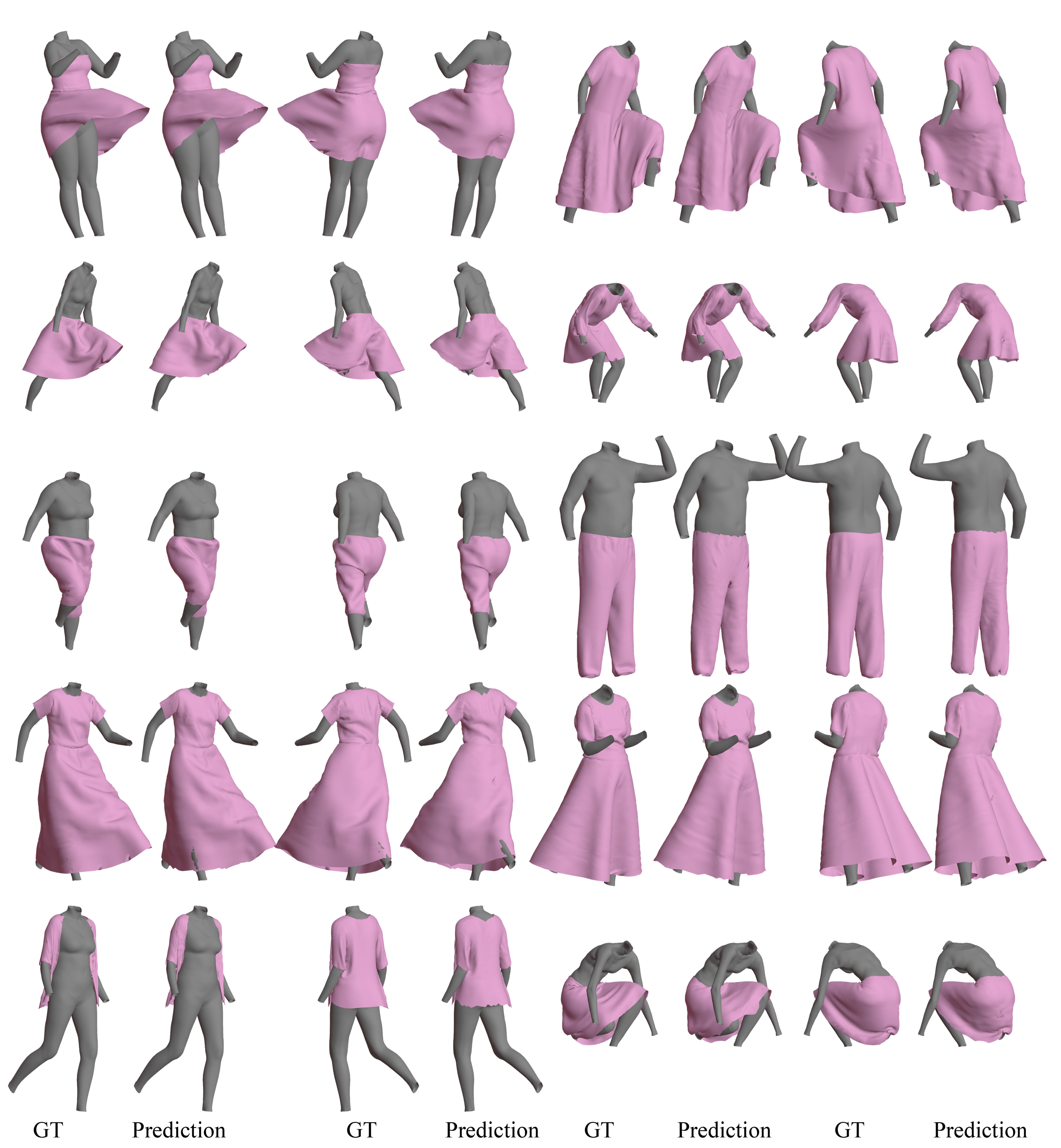}
   \caption{Additional qualitative examples from the CLOTH3D dataset: ground truth and proposed model's reconstruction.}
   \label{fig:examples_2}
\end{figure*}

\begin{figure*}[]
  \centering
   \includegraphics[width=0.95\linewidth]{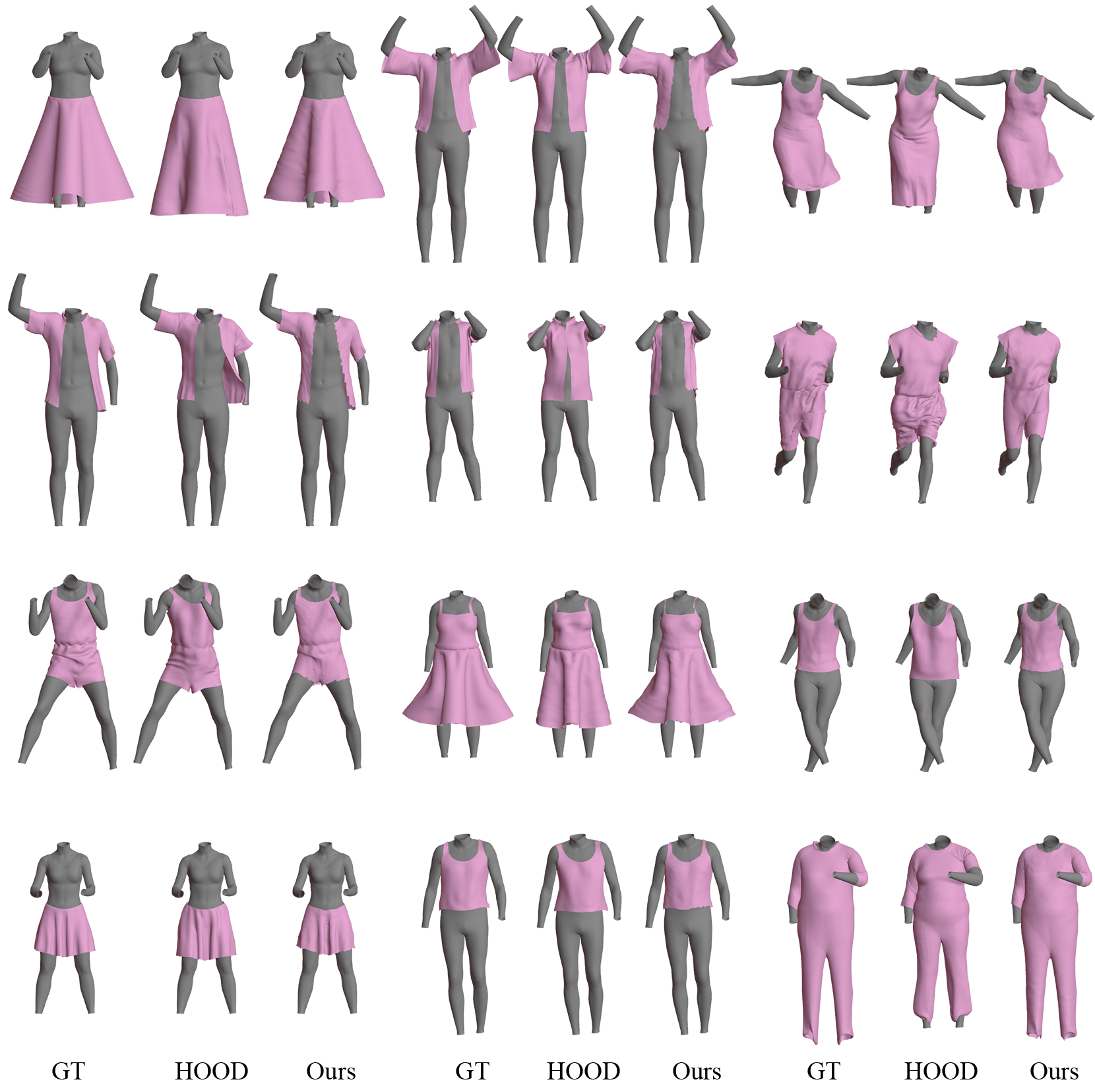}
   \caption{Additional qualitative comparison examples with HOOD, where we can see the ground truth, HOOD prediction, and ours.}
   \label{fig:hood_suppl}
\end{figure*}

We provide some additional qualitative examples from the CLOTH3D dataset in \cref{fig:examples_1} and \cref{fig:examples_2}. Our model is able to reconstruct a high level of detail on a variety of different garments. We also show some additional comparison with HOOD in \cref{fig:hood_suppl}.

\end{document}